\documentclass{article}
\usepackage{spconf,amsmath,epsfig}
\usepackage{booktabs}
\usepackage{xcolor, colortbl}
\usepackage{subfig}
\usepackage{makecell}
\usepackage{multirow}

\definecolor{Gray}{gray}{0.7} 
\colorlet{lowopacitygray}{Gray!15} 
\definecolor{mygreen}{RGB}{0, 176, 80}

\let\OLDthebibliography\thebibliography
\renewcommand\thebibliography[1]{
  \OLDthebibliography{#1}
  \setlength{\parskip}{0pt}
  \setlength{\itemsep}{0pt plus 0.3ex}
}

\begin{document}\sloppy

\def\x{{\mathbf x}}
\def\L{{\cal L}}

\title{Chain-of-Thought Prompting for Demographic Inference with Large Multimodal Models}
%

\name{Yongsheng Yu$^{1}$ \hspace{12pt} Jiebo Luo$^{1}$}
\address{$^{1}$Department of Computer Science, University of Rochester \\ \texttt{yyu90@ur.rochester.edu; jluo@cs.rochester.edu}\\}

\maketitle

\begin{abstract}
Conventional demographic inference methods have predominantly operated under the supervision of accurately labeled data, yet struggle to adapt to shifting social landscapes and diverse cultural contexts, leading to narrow specialization and limited accuracy in applications. Recently, the emergence of large multimodal models (LMMs) has shown transformative potential across various research tasks, such as visual comprehension and description. In this study, we explore the application of LMMs to demographic inference and introduce a benchmark for both quantitative and qualitative evaluation. Our findings indicate that LMMs possess advantages in zero-shot learning, interpretability, and handling uncurated 'in-the-wild' inputs, albeit with a propensity for off-target predictions. To enhance LMM performance and achieve comparability with supervised learning baselines, we propose a Chain-of-Thought augmented prompting approach, which effectively mitigates the off-target prediction issue.
\end{abstract}

\section{Introduction}
\label{sec:intro}

Demographic inference~\cite{yeung2020face} involves analyzing population data based on characteristics like age~\cite{fu2008human,kuprashevich2023mivolo}, gender~\cite{abderrahmane2020hand}, and ethnicity~\cite{fairface,sarridis2023flac}. Essential in fields such as sociology, marketing, and public health, it helps identify societal trends, understand consumer behavior, and inform public policies. It is crucial for addressing issues such as aging populations and population migration, significantly impacting societal and economic strategies.

\begin{figure}[t]
    \centering
    \vspace{3mm}
    \includegraphics[width=0.8\linewidth]{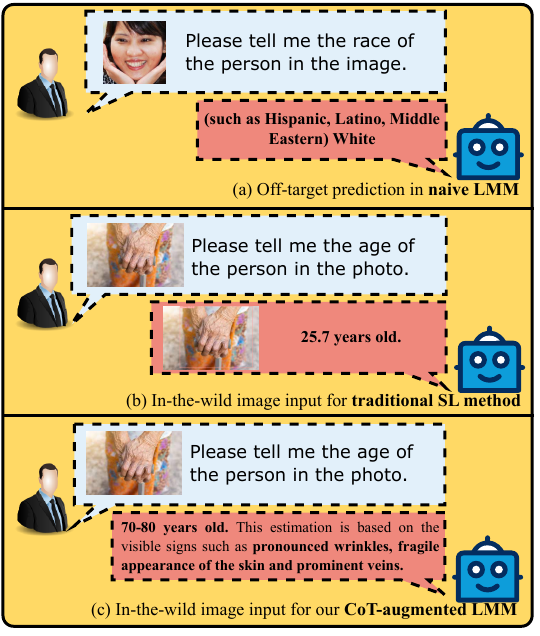}
    \vspace{-2mm}
    \caption{Analysis of traditional Supervised Learning (SL) methods and naive LMMs in demographic inference task.}
    \label{fig:feature}
\end{figure}

\begin{figure*}[t]
    \centering
    \includegraphics[width=\textwidth]{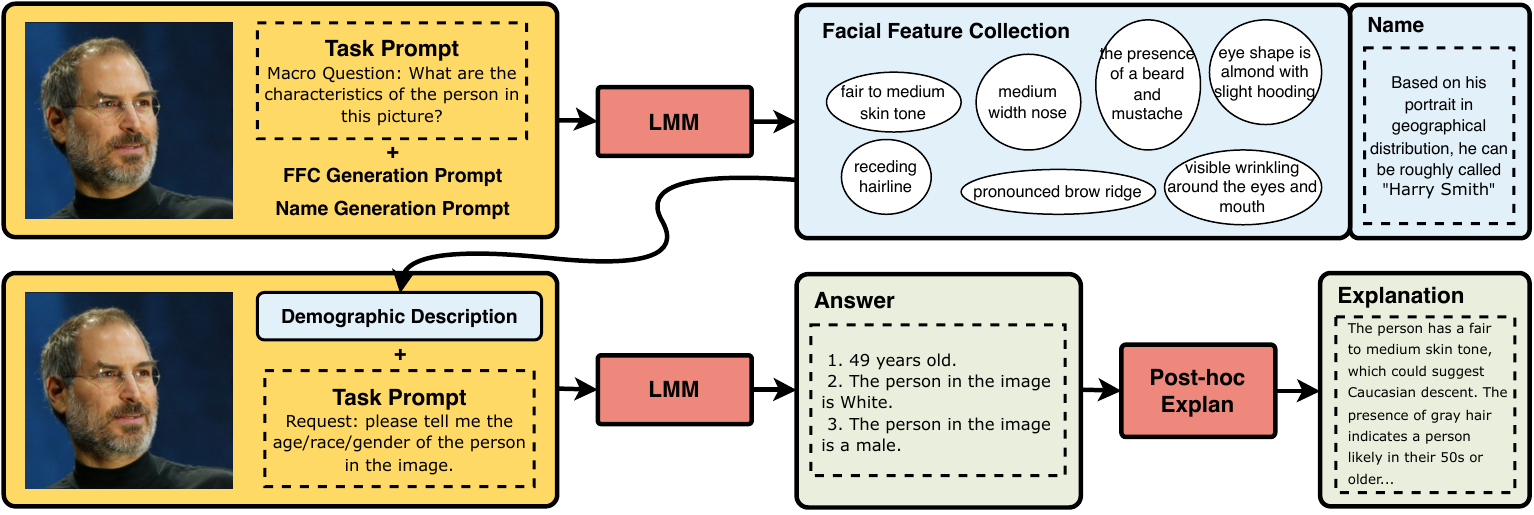}
    \caption{Conceptual workflow of our Chain-of-Thought prompting approach for demographic inference. The process begins with task prompts guiding LMM to articulate the facial features of the individual in the image, followed by name suggestions. Subsequently, the LMM employs these attributes as demographic descriptions to deduce age, race, and gender, and provides post-hoc explanations for its conclusions.}
    \label{fig:workflow}
\end{figure*}

Despite these advancements, current traditional artificial intelligence approaches to demographical inference typically rely on domain-specific, real-world labeled paired data. These methods are often limited in scope, lacking a comprehensive understanding of the Knowledge of Demographic Data, diversity among individuals and groups, cultural backgrounds, and macro socio-economic contexts. As a result, they fail to accurately predict demographic information in in-the-wild data and do not offer an interpretable process or suggestions for demographic inference.

Recently, the emergence of AI foundational models, led by Large Language Models (LLMs)~\cite{llama}, has provided a new paradigm. Characterized by their massive parameter count, training on broad and diverse data, and exceptional versatility, these models can adapt to a wide range of tasks through further training, such as evolving into large multimodal models (LMMs)~\cite{llava,chen2023minigpt} with the training of visual encoding models. This approach, capable of understanding and processing both image and text inputs, introduces a new paradigm in methodological design across various research fields.

In this study, we propose an integrated demographic inference benchmark and evaluate it on a series of popular open-source LMMs, namely LLaVA~\cite{llava}, MiniGPTv2~\cite{chen2023minigpt}, InstructBLIP~\cite{instructblip}, and internLM~\cite{zhang2023internlm}. These LMMs have already demonstrated impressive capabilities in understanding visual content and answering questions with a broad common-sense understanding and high natural language proficiency. Our investigation confirms that LMMs possess three main advantages over traditional demographic inference methods:
\begin{itemize}
    \item [1.] \textbf{Interpretability.} While traditional deep learning approaches yield results conforming to the format of training data labels, LMMs can easily allow the model to explain its predictions through post-hoc questioning.
    \item [2.] \textbf{Zero-shot prediction.} In our study, LMMs do not require any real-world labeled data for downstream task fine-tuning or few-shot data for context-providing instruction and can be directly applied to visual-based demographic inference test datasets.
    \item [3.] \textbf{Proficiency in handling out-of-domain data.} Traditional methods, trained on domain-specific data like cropped standard facial images, struggle with visual inputs like half-body portraits. In contrast, LMMs can accurately predict demographics in in-the-wild images, see Figure~\ref{fig:feature}(b) and (c).
\end{itemize}

Nevertheless, in in-domain test datasets, we have observed a significant gap between the performance of naive LMMs in a zero-shot setting and traditional supervised learning approaches. Additionally, the high degree of response flexibility in the language model of LMMs can lead to off-target predictions (see Fig.~\ref{fig:feature}(a)), a common issue when replacing traditional supervised models with LMMs for predictions. To address this, we also propose a Chain-of-Thought approach, enhancing LMMs' prompting with a two-step intermediary questioning process to obtain demographic feature descriptions. Our main contributions are as follows:
\begin{itemize}
    \item We are the first, to the best of our knowledge, to incorporate LMMs for demographic inference.
    \item We propose an integrated benchmark to assess the performance of models in demographic inference and study the effectiveness of LMMs on in-the-wild data.
    \item We introduce a Chain-of-Thought strategy to improve the performance of LMMs on demographic inference tasks while reducing the rate of off-target predictions.
\end{itemize}

\begin{figure*}[t]
    \centering
    \includegraphics[width=\textwidth]{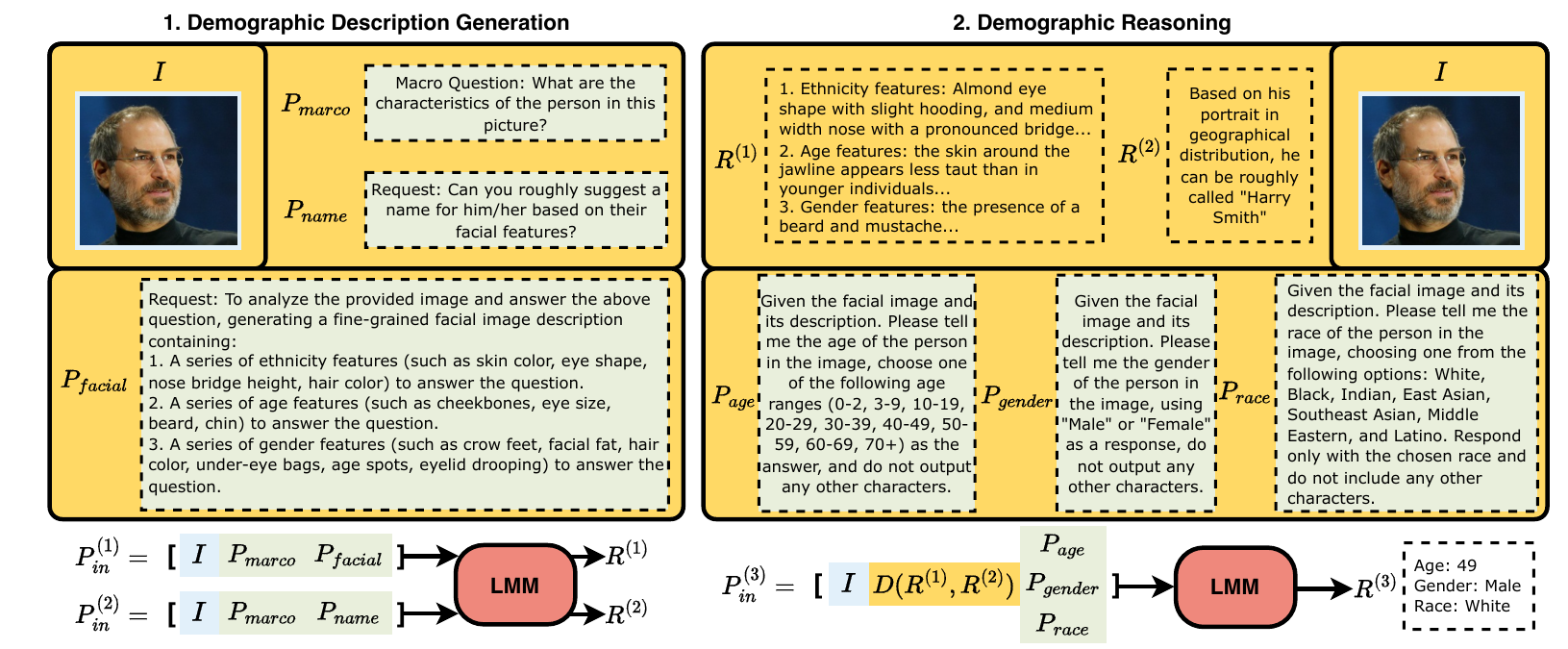}
    \caption{Full prompt example of the Chain-of-Thought augmented prompting for demographic inference. }
    \label{fig:cot-method}
\end{figure*}

\section{Related Works}
\label{sec:related}

\subsection{Demographic Inference}

In the evolving landscape of demographic inference, particularly in race, age, and gender prediction, recent advancements have been marked by the integration of deep learning. 
Gender prediction has similarly benefited from deep learning models~\cite{abderrahmane2020hand} applied to ocular and real-world images. 
In age prediction, the use of discriminant subspace learning~\cite{fu2008human} extracts aging patterns from facial images, presenting them as distinct manifold structures, thereby enhancing both age estimation and our visual understanding of the aging process. Deep learning systems, especially those trained on time-variant features~\cite{abderrahmane2020hand}, have shown remarkable results in age prediction from text and images. 
In this study, our primary focus is on the analysis and interpretation of three key demographic attributes: gender, age, and race.

\subsection{Large Multimodal Models}

Large Multimodal Models (LMMs)~\cite{llava,instructblip} represent a leap in AI technology, synthesizing the nuanced reasoning capabilities of Large Language Models (LLMs)~\cite{llama} with the perceptual insights of vision-language (VL) models. 
The transformative aspect of LMMs lies in their ability to perform sophisticated tasks that require concurrent understanding and generation of multi-modal data, such as in visual question answering~\cite{DBLP:journals/corr/abs-2401-02582} and social media understanding~\cite{DBLP:journals/corr/abs-2311-07547}. 
Our research diverges from traditional LMM applications, pioneering the use of these models for demographic inference. Unlike prior works which primarily leveraged LMMs for tasks like visual question answering or object recognition, our approach explores the untapped potential of LMMs in deciphering demographic information.

\subsection{Chain-of-Thought in LLMs}
The advent of Chain-of-Thought (CoT) prompting in Large Language Models (LLMs) has emerged as a significant development in AI, enhancing model reasoning capabilities. CoT enables models to process and articulate intermediate steps or 'thoughts' during problem-solving, akin to a human-like reasoning process. This method contrasts with traditional direct answer generation, offering a more transparent and interpretable approach. CoT, along with its extensions like self-consistency, Tree-of-Thought, and Graph-of-Thought, enriches the interaction between users and AI models, particularly in complex tasks requiring detailed explanations or step-by-step reasoning. The implementation of CoT in multimodal contexts further extends its utility, allowing for intricate inference across both textual and visual inputs. This approach demonstrates the evolving sophistication of AI models in mirroring human cognitive processes and their application in diverse problem-solving scenarios.

\begin{table*}[t]
    \centering
    \caption{Quantitative experimental results on the UTKFace dataset. \colorbox{lowopacitygray}{\strut Gray} row indicates supervised learning baselines.}
    \label{tab:utkface}
    \vspace{-2mm}
    \resizebox{\textwidth}{!}{
    \begin{tabular}{llllllllllc} \toprule
                 & \multicolumn{5}{c}{Age}         & \multicolumn{2}{c}{Gender} & \multicolumn{2}{c}{Ethnicity} &  \\ \cmidrule(r){2-6} \cmidrule(r){7-8} \cmidrule(r){9-10}
                 & MSE & RMSE & MAE & R$^2$ & MAPE & Accuracy   & Kappa  & Accuracy        & Kappa  & \multirow{2}{*}[2.5ex]{\makecell{Off-target \\ Rate}}     \\ \midrule \rowcolor{lowopacitygray}
    \textcolor{gray}{FLAC~\cite{sarridis2023flac}}     &  \textcolor{gray}{-} & \textcolor{gray}{-} & \textcolor{gray}{-} & \textcolor{gray}{-} & \textcolor{gray}{-} & \textcolor{gray}{-} & \textcolor{gray}{-} & \textcolor{gray}{0.9200}  &   \textcolor{gray}{-} & \textcolor{gray}{0.0\%} \\ \rowcolor{lowopacitygray}
    \textcolor{gray}{MWR~\cite{shin2022moving}}   &  \textcolor{gray}{-} & \textcolor{gray}{-} &  \textcolor{gray}{4.37} & \textcolor{gray}{-} &   \textcolor{gray}{-} & \textcolor{gray}{-} & \textcolor{gray}{-} & \textcolor{gray}{-} & \textcolor{gray}{-} & \textcolor{gray}{0.0\%}   \\ \midrule
    MiniGPTv2~\cite{chen2023minigpt}    &  132.25 & 11.50 & 7.28 & 0.6600 & 103.36\% & 0.9540 & 0.9071 &  0.5920 & 0.4656 & 7.9\%      \\
    InstructBLIP~\cite{instructblip} &   241.39 & 15.54 & 8.35 & 0.3794 & 232.51\% & 0.8980 & 0.7914 & 0.6140 & 0.4139 & 20.2\%      \\
    LLaVA~\cite{llava15}        &   60.14 & 7.75 & 5.13 & 0.8454 & 24.03\% & 0.9620 & 0.9236 &  0.8560 & 0.8005 & 0.2\%  \\ \midrule 
    LLaVA w/ COT      & 55.04 & 7.42 & 4.80 & 0.8585 & 15.35\% & 0.9750 & 0.9496 & 0.8810 & 0.8358 & 0.0\%  \\ \bottomrule
    \end{tabular} 
    }
\vspace{-2mm}
\end{table*}

\begin{figure*}[t]
    \centering
    \includegraphics[width=\textwidth]{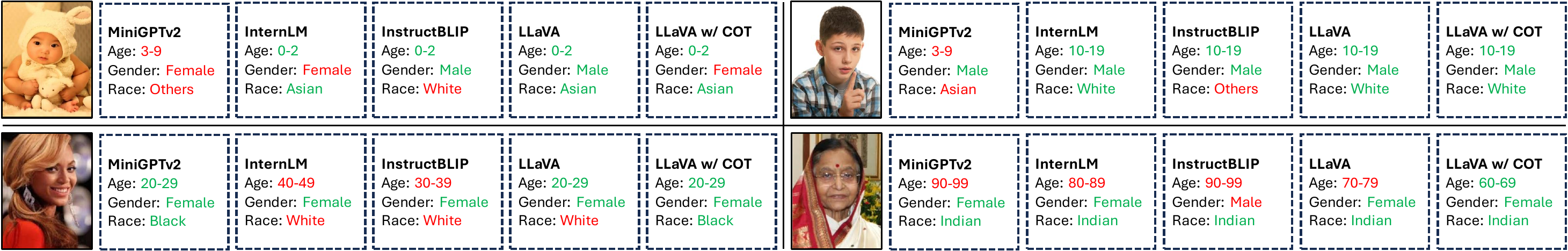}
    \caption{Qualitative comparison of naive LMMs and COT-augmented LMMs. \textcolor{red}{Red} answers are incorrect, \textcolor{mygreen}{green} ones are correct. 'Others'~\cite{utkface} in race categories includes those not White, Black, Asian, or Indian. Zoom in for a better view.}
    \label{fig:qual}
    \vspace{-0.1in}
\end{figure*}

\section{Methodology}

\subsection{Problem Formulation}
Our task can be conceptualized as a text generation problem under multimodal prompting. The core input to our model is a single-person photo featuring a face, upon which our demographical models infer demographic information such as gender, age, and ethnicity implicitly from visual features, based on textual task instructions. We assess the method's performance using distance metrics for paired data that are either continuous or discrete, aligning with the dataset's label data type. For dataset and experimental settings, please refer to the subsequent sections.

\subsection{Large Multimodal Models}
Based on text dialogues in large language models $f(\cdot)$ parameterized by $\theta$, LMMs are multimodal models that incorporate both visual and linguistic modalities as inputs. Typically, they receive a collection of images $I$ and a related textual task prompt $P$, and the large language model, based on the task prompt and visual input, can infer and respond in text form. More specifically, to process visual inputs, language models first map each modality to a shared representational space using pre-trained encoding models, i.e., the visual encoding model embeds $I$ into the textual space, resulting in $\phi(I)$, which is then combined with tokenized language embeddings $\tau(P)$ and fed into the large language model for textual response $R$.
\begin{equation}\label{eqn:init}
    R = f(\phi(I), \tau(P) ; \theta).
\end{equation}

The submodules mentioned above may vary in network architecture, pretraining methods, and parameters across different LMMs, such as LLaVA~\cite{llava} and MiniGPTv2~\cite{chen2023minigpt}, but generally follow the outlined pipeline.

\subsection{Chain-of-thought Augmented Prompting}

We find that prompting LMMs with native task instructions does not achieve performance comparable to supervised methods, as shown in Table~\ref{tab:utkface}. This is attributed to the excessive versatility of LMMs in handling tasks without being fine-tuned for demographic prediction, where straightforward task instructions fail to fully leverage the prior of LMM in a zero-shot manner.

To maximize the inherent inferential potential of LMMs, we propose enhancing textual prompts with a Chain-of-Thought approach. Our first step is to have LMMs directly interpret the input image, generating a detailed Facial Feature Collection (FFC) $P_{\text{facial}}$, thus avoiding the need for ground-truth caption data. As illustrated in Figure~\ref{fig:cot-method}, $P_{\text{facial}}$ systematically constructs key attributes about age, gender, and ethnicity, e.g., gender-related attributes might include crow's feet, facial fat, hair color, under-eye bags, and age spots.
\begin{equation}
    R^{(1)} = f(\phi(I), \tau(P_{\text{marco}}), \tau(P_{\text{facial}}) ; \theta).
\end{equation}

Ethnicity is a more challenging category to discern, as predictions rely not only on facial features, especially for an attribute like ethnicity with strong multicultural aspects and common knowledge dependencies (like birth geographical location, and nationality). In the second step, inspired by~\cite{yeung2020face}, we consider the person's name as a key feature for discerning ethnicity. Using the naming (captioning) capability of LMMs, we predict the last and first names based on the input image. We combine the estimated name $P_{\text{name}}$ with facial feature collection $P_{\text{facial}}$ to form a demographic description $D$, as seen in Figure~\ref{fig:cot-method}, which is used to enhance the original textual prompt.
\begin{equation}
    R^{(2)} = f(\phi(I), \tau(P_{\text{marco}}), \tau( P_{\text{name}}) ; \theta).
\end{equation}

It is noteworthy that in the demographic description, the roles of attributes are not entirely orthogonal; for instance, gender-related attributes might also aid in age prediction. Therefore, in the third step, we integrate the entire demographic description into the prompt for each demographic subcategory prediction as contextual supplementation. Thus, we utilize the generated demographic description as intermediate Chain-of-Thought inference steps, explicitly enhancing the overall input prompt for response generation, as follows:

\begin{equation}
    P^{(3)}_{in} = [I, D(R^{(2)}, R^{(2)}), P_{\text{age} \mid \text{gender} \mid \text{race}}].
\end{equation}

Under Chain-of-Thought enhancement, we modify the previous formula~\ref{eqn:init}, and the final response text R generated by LMMs is as follows:

\begin{equation}
    R^{(3)} = f(P^{(3)}_{in}; \phi, \tau, \theta).
\end{equation}

This leverages the capability LMM of to represent images in high-dimensional spaces and their descriptive power in language modeling. Notably, our enhancement method requires no fine-tuning or few-shot in-context prompting, remaining entirely text-based and zero-shot, and does not require any annotated facial image captions.

\begin{table*}[t]
    \centering
    \caption{Comparative performance and off-target prediction mitigation on the FairFace dataset.}
    \label{tab:fairface}
    \vspace{-3mm}
    
    \subfloat[Quantitative results on FairFace. \colorbox{lowopacitygray}{\strut Gray} row indicates supervised learning baselines.]{
    \resizebox{0.60\textwidth}{!}{
    \begin{tabular}{lllllll} \toprule
                 & \multicolumn{2}{c}{Age} & \multicolumn{2}{c}{Gender} & \multicolumn{2}{c}{Ethnicity} \\ \cmidrule(r){2-3} \cmidrule(r){4-5} \cmidrule(r){6-7}
                 & Accuracy     & Kappa    & Accuracy       & Kappa       & Accuracy        & Kappa       \\ \midrule \rowcolor{lowopacitygray}
    \textcolor{gray}{FairFace~\cite{fairface}}   &  \textcolor{gray}{0.597}  & \textcolor{gray}{-} & \textcolor{gray}{0.942}  & \textcolor{gray}{-} & \textcolor{gray}{0.937} & \textcolor{gray}{-}      \\ \rowcolor{lowopacitygray}
    \textcolor{gray}{MiVOLO~\cite{kuprashevich2023mivolo}}   &  \textcolor{gray}{0.611} & \textcolor{gray}{-} & \textcolor{gray}{0.957} & \textcolor{gray}{-} & \textcolor{gray}{-} &  \textcolor{gray}{-}      \\ \midrule
    MiniGPTv2~\cite{chen2023minigpt}    &  0.316  &  0.175  &  0.925  & 0.849  &  0.472  &  0.373   \\
    InternLM~\cite{zhang2023internlm}     &  0.500    &  0.399 &  0.955  & 0.910  &  0.462  & 0.351  \\
    InstructBLIP~\cite{instructblip} &  0.291  &  0.153  &  0.874  &  0.744    &  0.539   &  0.449   \\
    LLaVA~\cite{llava15}        &  0.499 & 0.398  &  0.956 & 0.912 &  0.618  &  0.546  \\  \midrule
    LLaVA w/ COT      &  0.577 & 0.490  &  0.958 & 0.916 &  0.692  &  0.634  \\ \bottomrule
    \end{tabular}
    }}
    \hfill 
    \subfloat[Mitigation of off-target prediction on FairFace via COT.]{
    \resizebox{0.38\textwidth}{!}{
    \begin{tabular}{lcc} \toprule
                        & Off-target Rate  & Accuracy \\ \midrule
    MiniGPTv2           & 18.4\%            & 0.316    \\
    MiniGPTv2 w/ COT    & 6.8\%              & 0.473    \\ \midrule
    InternLM            & 0.1\%             & 0.500    \\
    InternLM w/ COT     & 0.0\%             & 0.568    \\ \midrule
    InstructBLIP        & 17.1\%            & 0.291    \\
    InstructBLIP w/ COT & 5.5\%              & 0.434    \\ \midrule
    LLaVA               & 0.0\%             & 0.499    \\
    LLaVA w/ COT        & 0.0\%             & 0.577   \\ \bottomrule
    \end{tabular}}
    }
\vspace{-8mm}
\end{table*}

\begin{table}[]
    \caption{Quantitative experimental results on the CACD dataset. \colorbox{lowopacitygray}{\strut Gray} row indicates supervised learning baselines.}
    \label{tab:cacd}
    \resizebox{\linewidth}{!}{
    \begin{tabular}{lccccc}\toprule
                 & MSE & RMSE & MAE & R$^2$ & MAPE \\ \midrule \rowcolor{lowopacitygray}
    \textcolor{gray}{CORAL~\cite{cao2020rank}}   & \textcolor{gray}{-} & \textcolor{gray}{7.48} & \textcolor{gray}{5.25}  & \textcolor{gray}{-} & \textcolor{gray}{-}    \\ \rowcolor{lowopacitygray}
    \textcolor{gray}{MWR~\cite{shin2022moving}}   & \textcolor{gray}{-} & \textcolor{gray}{-} & \textcolor{gray}{4.37}  & \textcolor{gray}{-} & \textcolor{gray}{-} \\ \midrule
    MiniGPTv2~\cite{chen2023minigpt}    &  92.09   &   9.60   &  7.18     &  0.309   &   25.24\%   \\
    InternLM~\cite{zhang2023internlm}     & 83.44  & 9.13     &  7.39   & 0.374  & 25.18\%  \\
    InstructBLIP~\cite{instructblip}   &  78.44  &  8.86   & 6.27 &  0.412 & 21.85\% \\
    LLaVA~\cite{llava15}        &   76.19  & 8.73     &  6.65     &  0.428   &  22.51\%   \\ \midrule
    LLaVA w/ COT &   66.69 & 8.17 & 5.75 & 0.500 & 15.99\%   \\ \bottomrule
    \end{tabular}}
\end{table}

\section{Experiments}
\label{sec:exp}

This paper integrates a benchmark for evaluating LMMs in demographic inference, incorporating three facial image datasets: CACD~\cite{cacd}, FairFace~\cite{fairface}, and UTKFace~\cite{utkface}.
\begin{itemize}
    \item \textbf{CACD} comprises over 160,000 images of 2,000 celebrities, offering a broad spectrum in terms of age and appearance diversity. The ground truth age labels span from 14 to 54 years.

    \item \textbf{UTKFace} includes more than 20,000 facial images with age, gender, and ethnicity annotations. The age range is extensive, from 0 to 116 years. It encompass a wide array of subjects across different ethnicities and a balance of male and female participants.

    \item \textbf{Fairface} consists of 108,501 images, each labeled with age, gender, and race. It represents seven racial groups: White, Black, Indian, East Asian, Southeast Asian, Middle Eastern, and Latino.
\end{itemize}

Since there is no need to use these datasets' training sets to fine-tune LMMs for any downstream tasks, we select images from the official test splits of each dataset to serve as our benchmark for evaluating LMMs.

\subsection{Models}

On our benchmark, we evaluate four popular LMMs: LLaVA~\cite{llava}, InstructBLIP~\cite{instructblip}, InternLM~\cite{zhang2023internlm}, and MiniGPTv2~\cite{chen2023minigpt}. To implement these LMMs, we utilize their official code and pre-trained weights. Specifically, we select the 13B weights version of LLaVA 1.5~\cite{llava15}, the InstructBLIP based on vicuna 13B weights, MiniGPTv2 based on the llama-2 7B~\cite{llama} language model, and the 7B version of InternLM.

We apply our Chain-of-Thought enhancement to LLaVA. It's noteworthy that commercial models like GPT4V~\cite{gpt4v} and Gemini, which are not open-sourced and may update versions frequently, are not the focus of this paper regarding performance state-of-the-art (SOTA). Therefore, we only use the aforementioned open-source LMMs, ensuring experimental stability and reproducibility.

\subsection{Quantitative analysis}

We apply our proposed COT strategy to LLaVA, referred to as "LLaVA w/ COT". For the continuous age annotations in the UTKFace and CACD datasets, we utilize metrics such as MAE, MAPE, and RMSE for evaluation. For other discrete attributes, performance is measured using Accuracy and Kappa. As shown in Figure~\ref{tab:fairface}(a) and Figure~\ref{tab:utkface}, ``LLaVA w/ COT" exhibits top performance on both UTKFace and FairFace datasets, particularly in achieving $0.00\%$ off-target rate and high Kappa scores, indicating precise and reliable predictions across age, gender, and ethnicity categories, and significantly outperforming naive LMMs. As shown in Figure~\ref{tab:cacd}, ``LLaVA w/ COT" variant outperforms LMM baselines in the quantitative analysis on the CACD dataset, exhibiting the lowest Mean Absolute Error (MAE) at $5.75$ and the highest R$^2$ value of $0.500$, indicating its strong predictive accuracy and its ability to explain half of the variance in the dataset. Beyond LMMs, our method exhibits comparable performance to traditional supervised learning approaches. In Table~\ref{tab:fairface}(a), "LLaVA w/ COT" outperforms state-of-the-art supervised learning methods in gender prediction accuracy.

\subsection{Qualitative analysis}

In a qualitative comparison, CoT-augmented LMMs show a improvement in demographic prediction accuracy over naive LMMs. As shown in Figure~\ref{fig:qual}, it is evident in the consistent green (correct) labels across the augmented models, indicating precise age, gender, and race identification. The naive LMMs, however, frequently mislabel, particularly in the 'Others' race category, highlighting the difficulty of handling diverse demographic attributes without augmentation. Enhanced models display a higher degree of specificity and sensitivity to visual cues. An interesting observation is that utilizing CoT-augmented prompting for inference may occasionally be counterproductive when dependent on a singe image. For instance, in the top-left case of Figure~\ref{fig:qual}, the CoT-augmented prediction identifies the subject as female, potentially due to the interpretation of clothing details, such as the cute bunny-style outfit, which might be stereotypically associated with female infants. However, the ground truth is male.

\subsection{Off-target prediction}

Due to the high flexibility of the language model of LMMs, texts generated by them do not necessarily fit the ground truth categories, even with explicit prompting (as shown in Figure~\ref{fig:cot-method} with instruction ``not to output any other characters"). This off-target phenomenon occasionally enhances the readability of responses or provides some interpretable information, which is acceptable in contexts that do not require quantitative evaluation. However, it also results in ambiguous or irrelevant answers, as seen in Figure~\ref{fig:feature}(a). Therefore, we consider off-target prediction in demographic inference as an issue with LMMs. To fairly assess LMMs, for each demographic inference, if the output is off-target, the inference is repeated until an on-target prediction occurs or $N$ iterations have passed. As shown in Table~\ref{tab:fairface}(b), some LMMs still exhibit a high off-target rate despite this procedure. After $N$ repetitions, if the prediction remains off-target, we use the CLIP~\cite{radford2021learning} to encode prediction and measure the similarity between prediction and ground truth categories, selecting the highest similarity as a replacement. We set $N$ to $5$ in our experiments. Importantly, we find that the proposed COT method effectively mitigates off-target prediction issues. As depicted in Figure~\ref{tab:fairface}(b), MiniGPTv2 with COT sees a reduction in off-target rate from $18.4\%$ to $6.8\%$, and InstructBLIP with COT drops from $17.1\%$ to $5.5\%$. Both InternLM and LLaVA achieve a $0.0\%$ off-target rate when combined with COT, underscoring the efficacy of our COT in enhancing model precision.

\section{Conclusion}

Our study presents a comprehensive exploration of LMMs for demographic inference, establishing a benchmark for the field. We demonstrate that LMMs, when enhanced with a Chain-of-Thought prompting strategy, not only provide interpretable, zero-shot predictions that excel in uncurated scenarios but also show promising results in adapting to the diverse and dynamic nature of demographic data. The introduction of this Chain-of-Thought approach significantly reduces off-target predictions, aligning LMM performance closely with that of traditional supervised learning methods, as evidenced by our rigorous quantitative and qualitative evaluations across multiple datasets. Our findings highlight the potential of LMMs to revolutionize demographic inference, offering a flexible and robust alternative to existing AI models. We believe that the integration of such AI models can greatly benefit societal-scale applications, from policy formulation to personalized services, by providing nuanced and accurate demographic analyses.

\bibliographystyle{IEEEbib}
\bibliography{icme2023template}

\end{document}